 \documentclass[tablecaption=bottom,wcp]{jmlr} 

\usepackage{booktabs}
\usepackage{todonotes}
\usepackage{comment}
\usepackage{graphicx}
\usepackage{fancyhdr}

\usepackage[load-configurations=version-1]{siunitx} 

\theorembodyfont{\upshape}
\theoremheaderfont{\scshape}
\theorempostheader{:}
\theoremsep{\newline}

\newcommand{\mb}{\mathbf}
\newcommand{\Q}{Q_{\theta}}
\newcommand{\xb}{\mathbf{x}_{\mathbf{b}}}
\newcommand{\Xb}{\mathbf{X}_{\mathbf{b}}}
\newcommand{\pmax}{p_{\mathrm{max}}}

\jmlrproceedings{AABI 2019}{2nd Symposium on Advances in Approximate Bayesian Inference, 2019}

\title[Masking schemes for universal marginalisers]{Masking schemes for universal marginalisers}

\author{\Name{Anonymous Authors}\\
  \addr Anonymous Institution}

  \author{\Name{D. Gautam} \Email{divya.gautam@babylonhealth.com}\\
   \Name{M. Lomeli} \Email{maria.lomeli@babylonhealth.com}\\
   \Name{K. Gourgoulias} \Email{kostis.gourgoulias@babylonhealth.com}\\
   \Name{D.H. Thompson} \Email{daniel.thompson@babylonhealth.com}\\
   \Name{S. Johri} \Email{saurabh.johri@babylonhealth.com}\\
   \addr Babylon Health, London, UK}

\begin{document}
\maketitle

\begin{abstract}
We consider the effect of structure-agnostic and structure-dependent masking schemes when training a universal marginaliser~\citep{Douetal17} in order to learn conditional distributions of the form $P(x_i |\mb x_{\mb b})$, where $x_i$ is a given random variable and $\mb x_{\mb b}$ is some arbitrary subset of all random variables of the generative model of interest. In other words, we mimic the self-supervised training of a denoising autoencoder, where a dataset of unlabelled data is used as partially observed input and the neural approximator is optimised to minimise reconstruction loss. We focus on studying the underlying process of the partially observed data---how good is the neural approximator at learning all conditional distributions when the observation process at prediction time differs from the masking process during training? We compare networks trained with different masking schemes in terms of their predictive performance and generalisation properties.

\end{abstract}

\section{Motivation}
In automated medical diagnosis a Bayesian network (BN) can be used as a statistical model for risk factors, diseases and symptoms in order to provide medical recommendations. For a BN to be appropriate for medical diagnosis, it typically needs to capture the relationships between thousands of diseases, symptoms and risk factors, making exact inference intractable. In this work, we focus on unlabeled binary data generated from a three-layer risk-disease-symptom BN as the generative model of interest.

During a typical diagnosis, only a subset of the evidence will be available, and hence we need to be able to efficiently (implicitly) marginalise over unobserved symptoms and risk factors. To obtain estimators of the posterior marginal distributions of diseases, we follow the amortised inference paradigm~\citep{GerGoo14} and train a denoising autoencoder~\citep{Vinetal10, Benetal13} using a variety of masking schemes for the input. This results in neural approximators for arbitrary conditional probabilities~\citep{Douetal17,Beletal19}.
In this scenario, the quantities of interest are the posterior probability of diseases given sets of evidence containing observations from risk factors and/or symptoms from each patient.
These estimates can then be used for diagnostic accuracy, namely, how high the algorithm ranks the reference diagnoses on a differential diagnosis~\citep{Shwetal91dia}. Since the goal is to provide medically accurate recommendations, it is crucial to obtain accurate marginal estimates as well as to have computationally feasible inference schemes given the exponential number of queries required.

The main contribution of this work is a study of the effect on learning conditional distributions when the pattern of observed values in the data is misspecified. We consider multiple masking distributions, some of which are independent of the underlying structure of the BN, and others which are more tailored to it. 

\section{Background and related work}\label{sec:def}

Let $\mb X=\{X_1,\ldots, X_n\}$, $X\in \{0,1\}^n$, denote the collection of random variables in a BN of $n$ nodes with joint distribution denoted by $P(\mb x)$. $M(\mb b)$ is the masking distribution for the masking random variable $\mb B\in \{0,1\}^n$. The masks are applied to the nodes through the element-wise product: $\mb x* \mb b=(x_1b_1,\ldots, x_nb_n)$. If $b_i=1$, we say that the value $x_i$ is observed, otherwise it is unobserved or masked. We will also use the notation $\xb$ to denote the subset of unmasked nodes.

We use the universal marginaliser (UM)~\citep{Douetal17}, a feedforward neural network that takes masked samples $\mb x*\mb b$ from a given BN as input, and learns to predict the marginals of all the nodes in the BN conditioned on the assignment of the observed nodes in the input. It is trained using the multi-label binary cross entropy loss:
\begin{align}
\label{eq:loss-function}
    &\min_{\theta}\left \{\mathbb{E}_{P(\mb x)}\mathbb{E}_{M(\mb b)}\sum_{i=1}^{n}CE(x_i|\mb x, \mb b; \theta)\right \},\text{where}\nonumber\\
    &CE(x_i|\mb x, \mb b; \theta)=-x_i\log \Q (X_i=1|\xb)-(1-x_i)\log (1-\Q(X_i=1|\xb))
\end{align}
in the same manner as a denoising autoencoder~\citep{Vinetal10}. As a consequence, the UM, denoted by $\Q$, learns to output $\Q(X_i=1|\xb)\approx P(X_i=1|\xb)$, for all $i=1,\ldots,n$. Joint posterior distributions can be estimated using the product rule and repeated evaluations of the UM. Section~\ref{app:nntrain} in the supplementary material contains further details about our implementation.

There exist other types of universal marginalisers that use a variety of generative probabilistic models based on other neural networks to estimate arbitrary conditional distributions. \citet{GerGreMurLar15} modify an autoencoder neural network to estimate conditional probabilities with autoregressive constraints. The authors mask the weighted connections of a standard autoencoder to convert it into a distribution estimator.
\citet{Ivaetal19} propose a variational autoencoder~\citep{KinDieWel14} with arbitrary conditioning. The authors use stochastic variational Bayes during the training together with the reparameterisation trick as well as a masking distribution. \citet{Beletal19} propose a generative adversarial network~\citep{Gooetal14} to learn every single conditional distribution: they use gradient based regularisation as well as a masking distribution during training. It is possible to apply the masking schemes proposed in section~\ref{sec:masks} to these other types of universal marginalisers. Masking the input layer is a special case of dropout~\citep{srivastava2014dropout,baldi2014dropout}, which is used to regularize over-parameterised neural networks by reducing the importance of individual neurons in the network during training.%
%


\section{Masking distributions}\label{sec:masks}

We consider a variety of masking distributions $M(\mb b)$ that, together with the generative model $P(\mb x)$, produce the data that we  use to train the UM as $\mb x * \mb b\sim M(\mb b)P(\mb x)$, where $*$ denotes element-wise product.

\noindent
\textbf{Uniform power setwise (uniform):} Masks are sampled uniformly from the power set of possible masks $\mb b\in \{0,1\}^n$.

\noindent
\textbf{Uniform sizewise (sizewise):} The distribution of the number of ones in $\mb b$ sampled using the uniform method above is $\text{Binomial}(n, 0.5)$, and hence the samples contain on average $n/2$ ones; see Figure~\ref{fig:my_gistogram} in the supplementary material for an illustration of this. If more samples from the tails of the Binomial  are desired, a distribution that selects $\mb b$ uniformly in terms of the number of ones $k$ therein can be achieved by first sampling a mask size $k \sim \text{Uniform}(\{0,\hdots,n\})$,
and then sampling a permutation of the array $(1,\hdots,1,0,\hdots,0)$ with $k$ ones and $n-k$ zeros.

%
\noindent
\textbf{Independent nodewise (nodewise):} Nodes are masked individually and independently rather than sampling an entire mask. Specifically, for each batch of samples, we sample $p\sim \text{Uniform}[0,\pmax]$ and set $M(b_i=1)=p$ for that batch, where $\pmax$ is a hyperparameter of the model. As with sizewise masking above, this procedure controls the number of evidence sets of different sizes seen during training: each sampled value of $p$ effectively specifies a maximum evidence set size for the masks in that batch. 
In this work we use $\pmax=1$ in order to sample from all possible evidence sets during training, but with a different distribution to the two methods described above.


\noindent
\textbf{Deterministic cycle (deterministic):}
This is a deterministic version of the nodewise method above: instead of sampling the probability $p$ of observing the nodes individually from a uniform distribution each batch, a list of values in $[0,1]$ is preselected, and we cycle through these values proceeding to the next one each batch. The rationale is to avoid clusters of $p$ values that can arise from sampling from the uniform distribution. Note that this method together with the above three are all \textbf{structure-agnostic} masking methods.


\noindent
\textbf{Markov blanket (markov):} For each mask, all nodes are initially masked, then one disease node is chosen at random, and the nodes in its Markov blanket are unmasked using the independent nodewise scheme above. The Markov blanket of a node $X$ is given by $\text{MB}(X)=\text{Pa}(X)\cup\text{Ch}(X)\cup\text{Pa}(\text{Ch}(X))$, where $\text{Pa}(X)$ and $\text{Ch}(X)$ correspond to the parents and children of $X$ respectively. This is an example of a \textbf{structure-dependent} masking method.

\section{Experiments}\label{sec:exp}

In this section, we compare the performance of the UM trained with loss function~\eqref{eq:loss-function} using the different masking distributions described in Section~\ref{sec:masks}.
The target (synthetic) Noisy-OR BN consists of $n=24$ nodes arranged in three layers of eight nodes each, representing risk factors, symptoms and diseases respectively. See Appendix~\ref{app:nntrain} for details about the architecture and training of the UM, and Appendix~\ref{app:noisy} for the structure of the target BN.

We tested the UM's ability to reproduce conditioned marginals $P(X|\Xb=\xb)$ for evidence sets $\{\Xb=\xb\}$ which were randomly chosen under two different observation models.
In the first case, we assume that evidence is observed under the uniform power setwise masking scheme.
In the second case, the test set construction uses the Markov masking assumptions---evidence is picked only from a randomly chosen disease node's Markov blanket.
The assignments of the unmasked nodes in both cases were chosen arbitrarily.
We constructed these test sets in order to capture the behaviour of each masking scheme when used for \emph{training} a UM under different \emph{prediction}-time observation patterns, mimicking the clinical diagnosis scenario.
\begin{figure}
    \centering
    \subfigure[Uniform evidence \label{fig:uni_masking_a} ]{\includegraphics[width=0.45\textwidth]{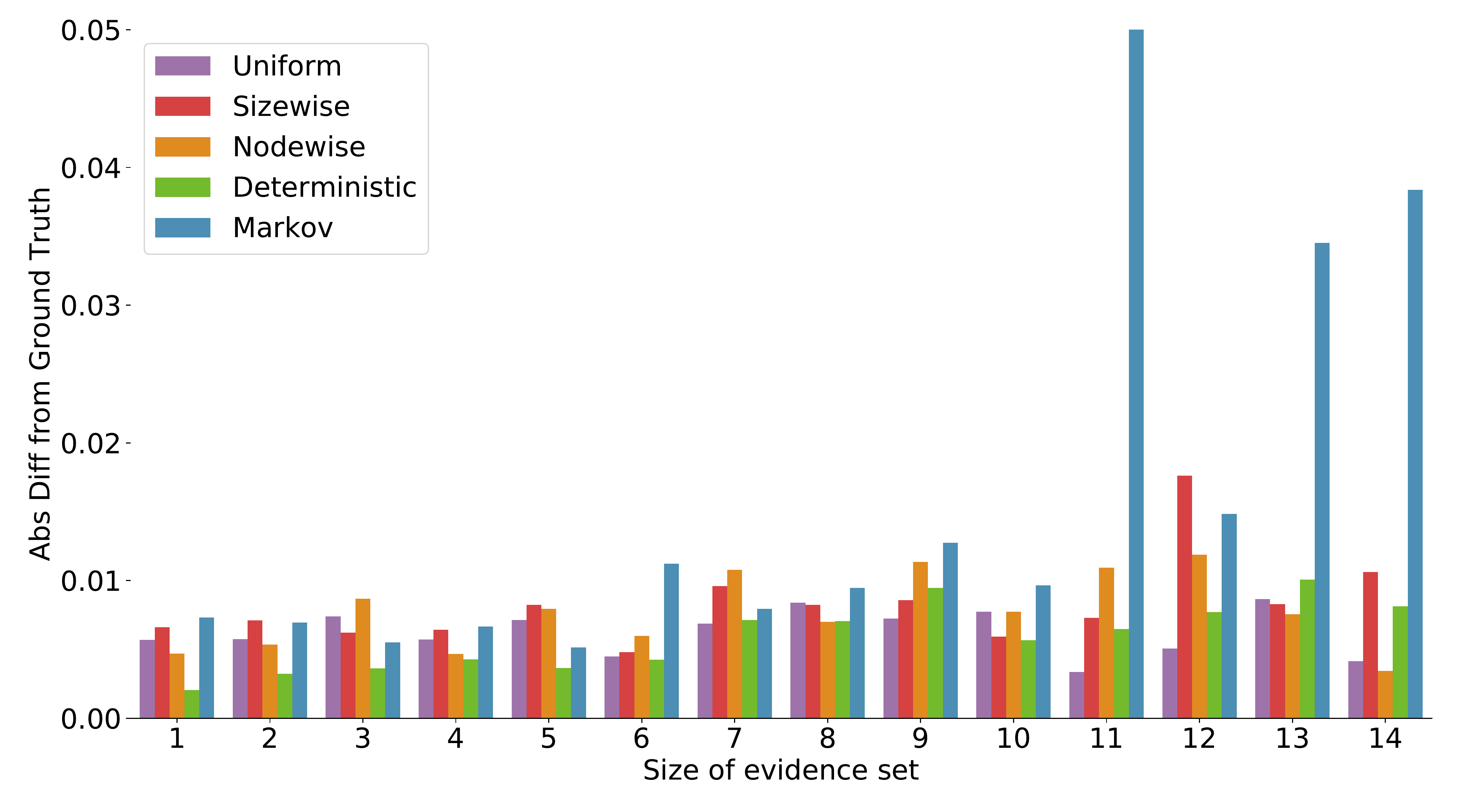}}
    \subfigure[Markov evidence
    \label{fig:uni_masking_b}
    ]{\includegraphics[width=0.45\textwidth]{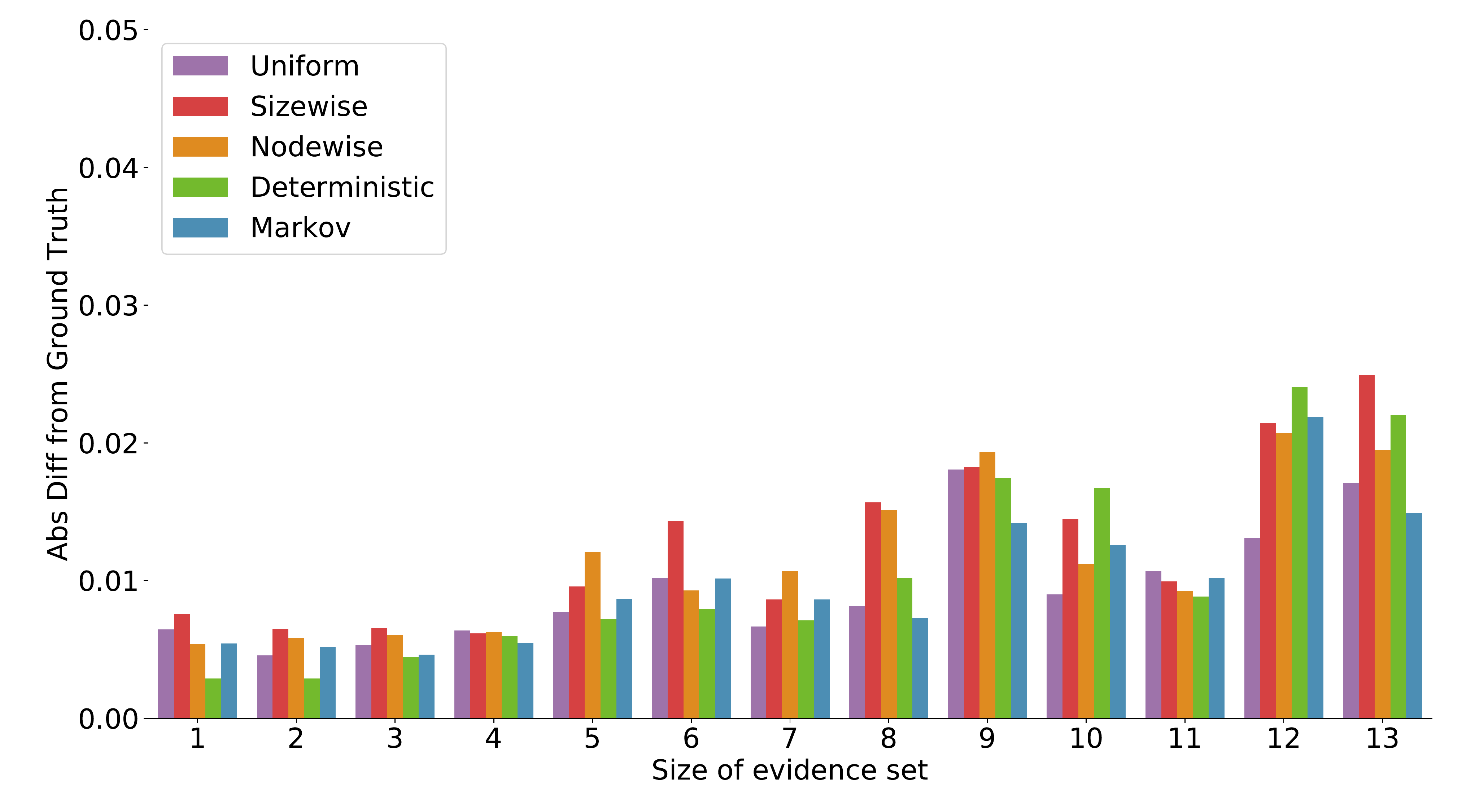}}
    \caption{\label{fig:error_masking_plot}
    Absolute difference $|\Q(X=0|\xb)-P(X=0|\xb)|$ between the UM's prediction and the ground truth as a function of evidence set size $k$ over $\{\xb\}$, where the states $\mb x$ are picked arbitrarily and $\mb b\sim M(\mb b )$, $M$ is either the uniform power setwise masking distribution or the Markov blanket one. The queried disease node $X$ is fixed across bars and plots.}
\end{figure}
Our comparison looks at the absolute difference $|\Q(X=0|\xb)-P(X=0|\xb)|$ between the UM's prediction and the ground truth for a given disease node $X$ as a function of evidence set size $k$. The ground truth $P(X=0|\xb)$ can be obtained in small Bayesian networks using exact inference methods.
In Figure~\ref{fig:uni_masking_a}, we see that in the case of uniform evidence all masking methods are competitive for small $k$, with deterministic masking being better on average in that regime.
Markov masking progressively gets worse as $k$ increases---the reason for this behaviour is that the Markov masking never chooses some of the masking patterns during training.
When we mask the evidence with respect to the Markov distribution (Figure~\ref{fig:uni_masking_b}), the Markov masking becomes competitive again across different $k$. 

The UM in Figure~\ref{fig:error_masking_plot} was trained with a large number of samples for each masking scheme, which makes the effect of the structure-agnostic masking distributions less apparent in the context of our small BN; see Appendix~\ref{app:add_exp} for examples of the UM trained for fewer epochs where differences are more pronounced.

We conjecture that using different masking schemes for training and corruption has an effect due to the inherent structure in the dataset, for example when using data from a Bayesian network or image data. This effect could potentially be less pronounced if the dataset does not contain any structure. In Appendix~\ref{app:add_exp}, we conducted an additional experiment to verify this claim empirically. We benchmarked some of our proposed masking schemes with a different universal marginaliser that uses a variational autoencoder architecture from~\citep{Ivaetal19}. The yeast dataset \citep{Dua:2019} was used, which is an example of independent and iid data and different masking and corruption schemes were picked.

\section{Conclusions}\label{sec:concl}
A number of structure-agnostic masking schemes are presented and its performance has been evaluated using a structured synthetic dataset from a Bayesian network. The main conclusion of this work is that the choice of masking scheme employed during training impacts both predictive performance and training efficiency, this choice should be informed by the quantities of interest at prediction time for maximal benefit. Model generalisation is also affected by the choice of masking method when the corruption process differs from the masking used during training or is unknown, as in real world scenarios.

A natural extension to this work is to benchmark the structure-dependent and structure-agnostic masking schemes on real-world clinical case data and larger Bayesian networks.

Finally, we have used our masking schemes for other universal marginalisers such as the variational autoencoder with arbitrary conditioning~\citep{Ivaetal19} to see the effect of masking schemes for iid data. Further experiments can be conducted with different neural network architectures such as the generative adversarial network from~\citet{Beletal19} and more complex datasets to study the performance of universal marginalisers for structured data. We believe this is an interesting avenue of future work.

%

\small
\bibliographystyle{apalike}
\bibliography{posterref}

\newpage

\appendix\label{sec:supp}

\section{Distribution of possible masks of a given size}

\begin{figure}[h]
    \centering
    \includegraphics[width=0.8\textwidth]{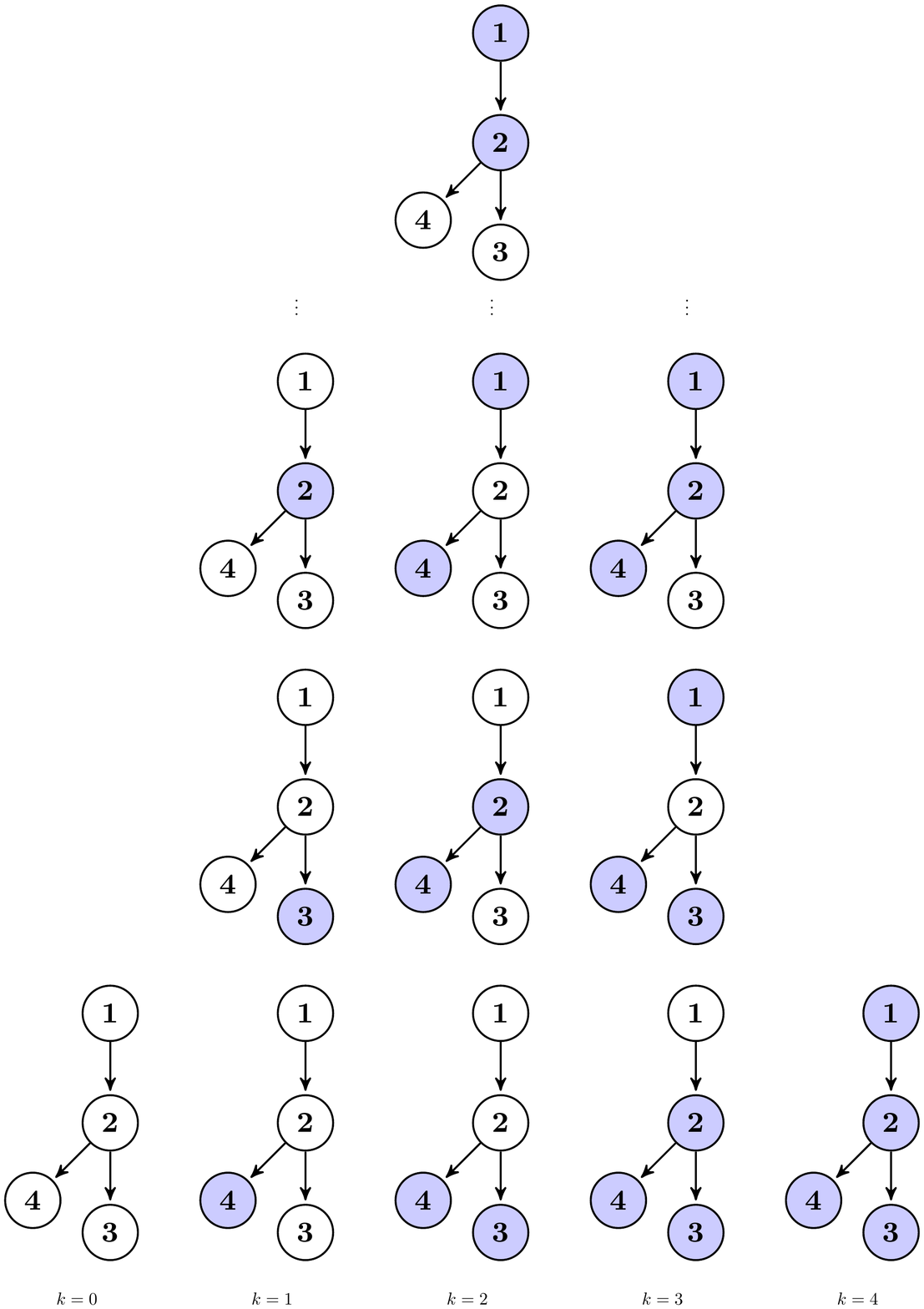}
    \caption{Illustration of the non-uniform distribution of the possible masks of size $k$ for $k\in \left\{0,\hdots,4\right\}$, for a BN of $n=4$ nodes.}
    \label{fig:my_gistogram}
\end{figure}

\section{Neural network architecture and training\label{app:nntrain}}

\begin{figure}
    \centering
    \includegraphics[width=0.5\textwidth]{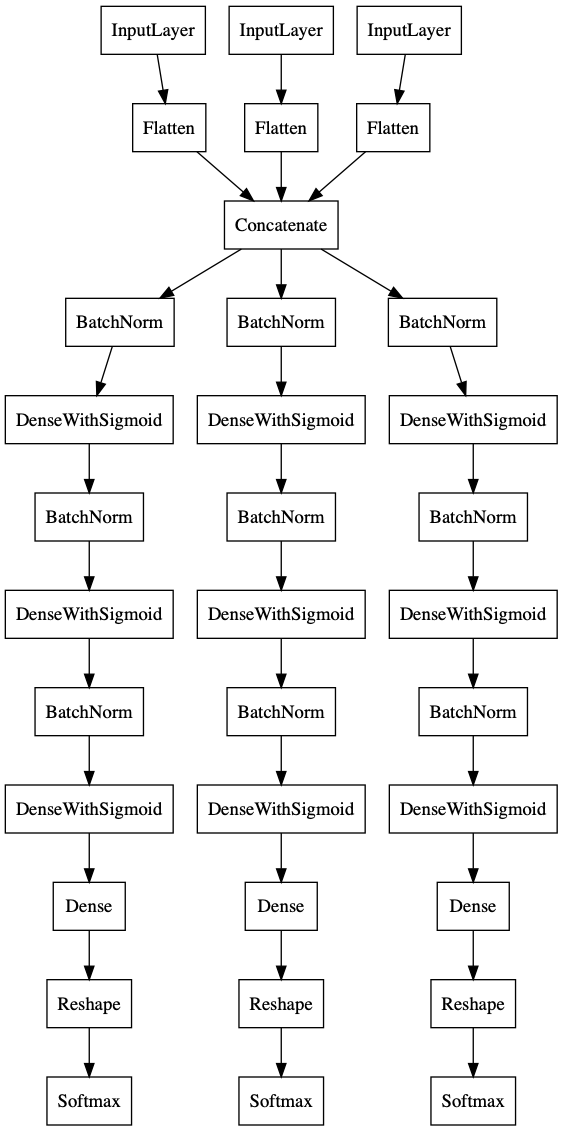}
    \caption{Neural network architecture. The three branches correspond to the $R$, $D$ and $S$ nodes of a three-layer $RDS$ BN.}
    \label{nnfig}
\end{figure}

Figure~\ref{nnfig} shows the neural network (NN) architecture, which consists of an input layer followed by a number of distinct multi-layer perceptron (MLP) branches, one for each layer of descendants in the BN, e.g.\ three branches for the $R$, $D$ and $S$ layers of a three-layer $RDS$ BN network.

Each MLP branch consists of 3 fully-connected layers of 512 hidden units each, to which batch normalisation~\citep{ioffe2015batch} before---and sigmoid activation after---is applied; followed by a final fully-connected layer of $N\times2$ hidden units, where $N\in\{N_R, N_D, N_S\}$, and $\{N_R, N_D, N_S\}$ correspond to the numbers of $R$, $D$ and $S$ nodes in the BN respectively.
Lastly softmax activation is applied on a per node basis.
This is done because of the 2 $\times$ Boolean parameterization of the node assignments, where $(F, F)$ corresponds to unobserved/masked, $(T, F)$ to observed and false, and $(F, T)$ to observed and true.
Hence the NN ends up learning two marginals for each node: one corresponding to the probability that the node is false, and another corresponding to the probability that it is true.
The softmax activation ensures these two probabilities sum to 1.

The NN is trained using the Adam optimizer~\citep{Kin14}, with the default parameter values and a learning rate of $10^{-4}$.

\section{Noisy-OR model\label{app:noisy}}

The number of parameters in a BN can be potentially large if the network is dense.
Namely, each node $i$ has $O(2^{|\text{Pa}_\mathcal{G}(i)|})$ parameters where $|\text{Pa}_\mathcal{G} (i)|$ is the number of parents of node $i$.
In order to have a more compact representation for its parameters, only storing $O(|\text{Pa}_\mathcal{G}(i)|)$ parameters per node, the Noisy OR model was proposed~\citep{Shwetal91}. 

The top layer of nodes corresponds to risk factors, the second layer, to disease nodes and the third layer, to symptom nodes.
The probability of a symptom node being false, given that only one disease is true and the rest are false, is  $P(S_j=0| \text{only}\ D_i =1) =\gamma_{i0}\times\gamma_{ij}$, where $\gamma_{i0}$ is the leak probability which corresponds to the case where all diseases are false.
Analogously, the probability of a disease being false, given that only one risk factor is true, is $P(D_i=0| \text{only}\ RF_r =1) =\lambda_{r0}\times \lambda_{ir}$, where $\lambda_{r0}$ is the leak probability for the case where all risk factors are false.
The prior for risk factor $r$ being false is denoted by $P(RF_r=0)=\pi_j$.
The general form for the probability of a risk factor, disease or symptom node being false given the values of its parents can be written in the following way:
\begin{align}
P(RF_{1} =r_1,\hdots,RF_{n}=r_n) = \prod_{j=1}^n&\pi_j^{1-r_j}(1-\pi_j)^{r_j}\nonumber\\
P(D_i=0| RF_{1} =r_1,\hdots,RF_{n}=r_n) &=\lambda_{i0} \prod_{j=1}^n\lambda_{ij}^{r_j} \nonumber\\
P(S_j=0| D_{1} =d_1,\hdots,D_{m}=d_m) &=\gamma_{j0} \prod_{k=1}^m\gamma_{jk}^{d_k} \nonumber.
\end{align}

The prior probabilities of diseases and risk factors can obtained from epidemiological data, where available. The conditional probabilities $P(S_j=0|\text{only }D_i=1)$ and $P(D_i=0| \text{only }RF_r =1)$ can be obtained through elicitation from multiple independent medical experts or from electronic medical records. The network structure expresses the corresponding clinical knowledge in graphical form, namely, whether or not there exists a direct relationship between a given pair of nodes..
See~\citet{ShwCoo91} for details about how to obtain the corresponding priors from QMR frequencies for a two layer BN and~\citet{Shwetal91}, for another method to obtain the corresponding conditional probabilities. \citet{BauKolSin97} and~\citet{TonKol00} propose an online parameter estimation and active learning methods respectively for parameter estimation in BNs.
\citet{Pop80} discusses structure finding methods for BNs.

\section{Additional Experiments}
\label{app:add_exp}
\subsection{Masking schemes with data from a Bayesian network}

\begin{figure}
    \centering
    \subfigure[Uniform evidence \label{app:fig:uni_masking_D2} ]{\includegraphics[width=0.45\textwidth]{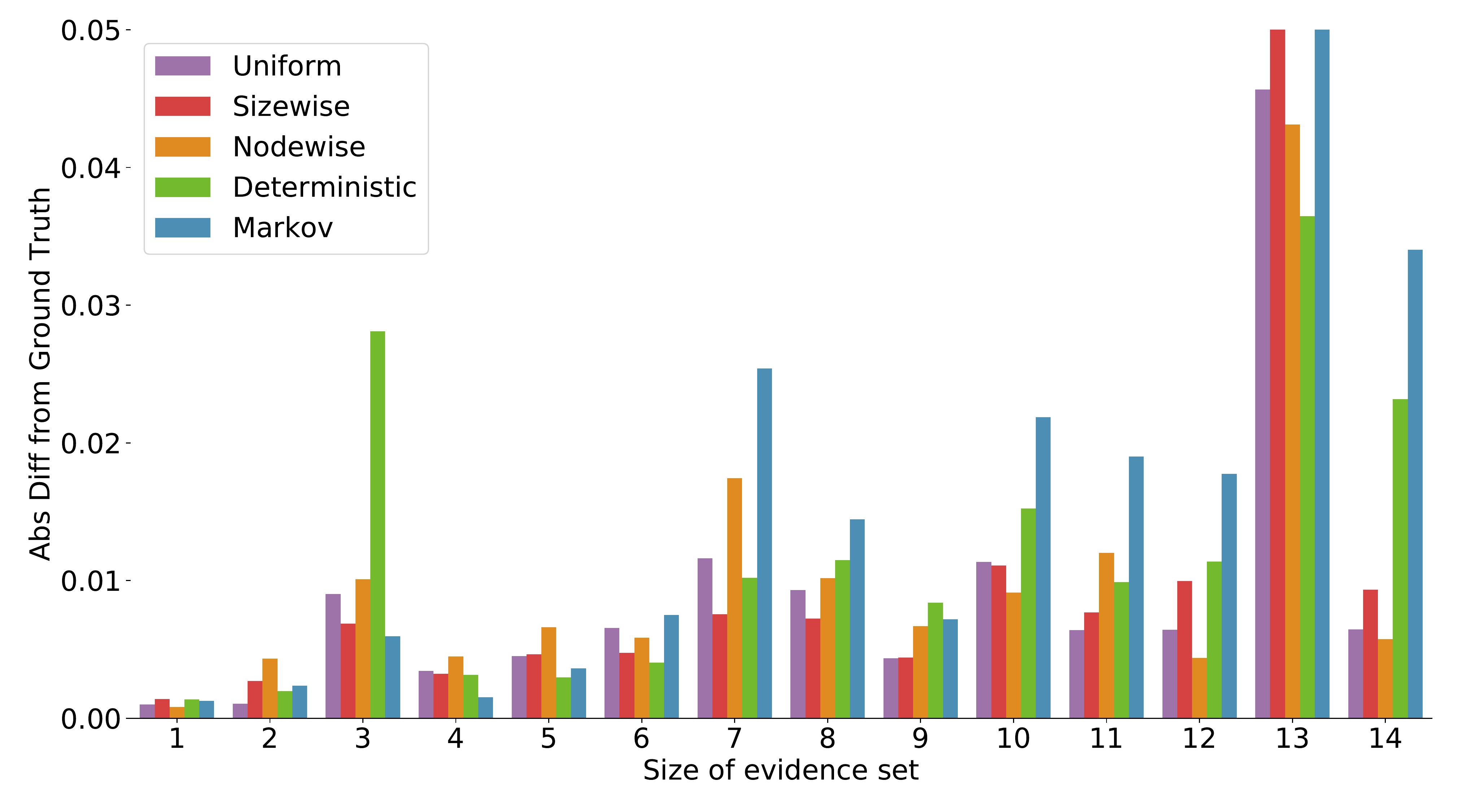}}
    \subfigure[Markov evidence
    \label{app:fig:markov_masking_D2}
    ]{\includegraphics[width=0.45\textwidth]{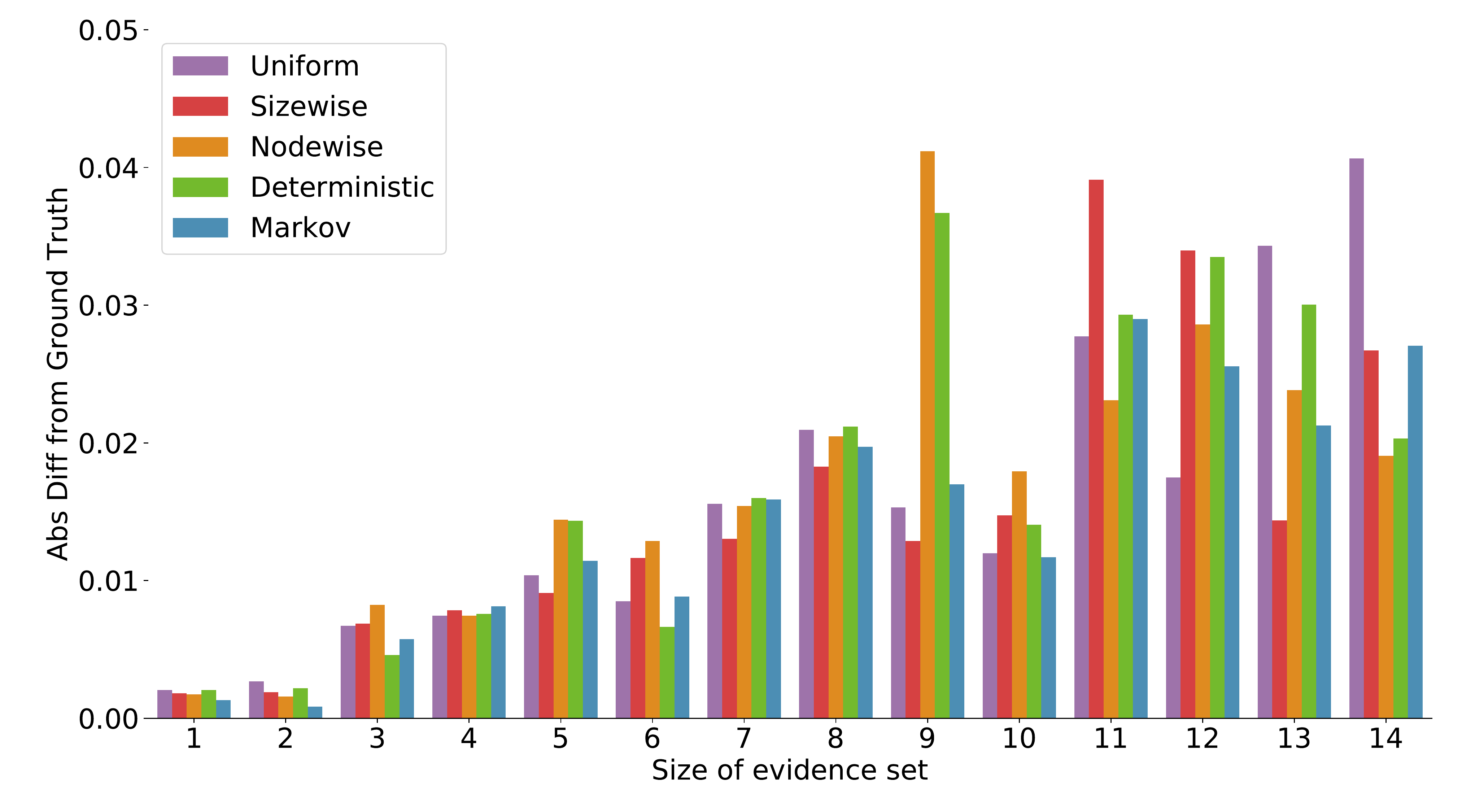}}
    \caption{\label{app:fig:D2_full_trained_error_plot}
    These plots correspond to a different query and are similar to Figure~\ref{fig:error_masking_plot}. Deterministic cycle masking is the best performing masking scheme for small evidence sets.}
\end{figure}

In this appendix, we share a few more details from our experiments.
Figure~\ref{app:fig:D2_full_trained_error_plot} is similar to the plot in the main paper, but for a different query disease node $X$.
In this example it is somewhat more apparent that the deterministic-cycle masking is the best performing masking scheme for small evidence sets.
In Figure~\ref{app:fig:uni_masking_D2}, the UM trained using Markov masking does not generalise well to uniformly-chosen evidence but is well-behaved over all the Markov-chosen evidence set sizes for which it was trained (Figure~\ref{app:fig:markov_masking_D2}).

\begin{figure}
    \centering
    \subfigure[Uniform evidence \label{app:fig:uni_masking_D1_short} ]{\includegraphics[width=0.45\textwidth]{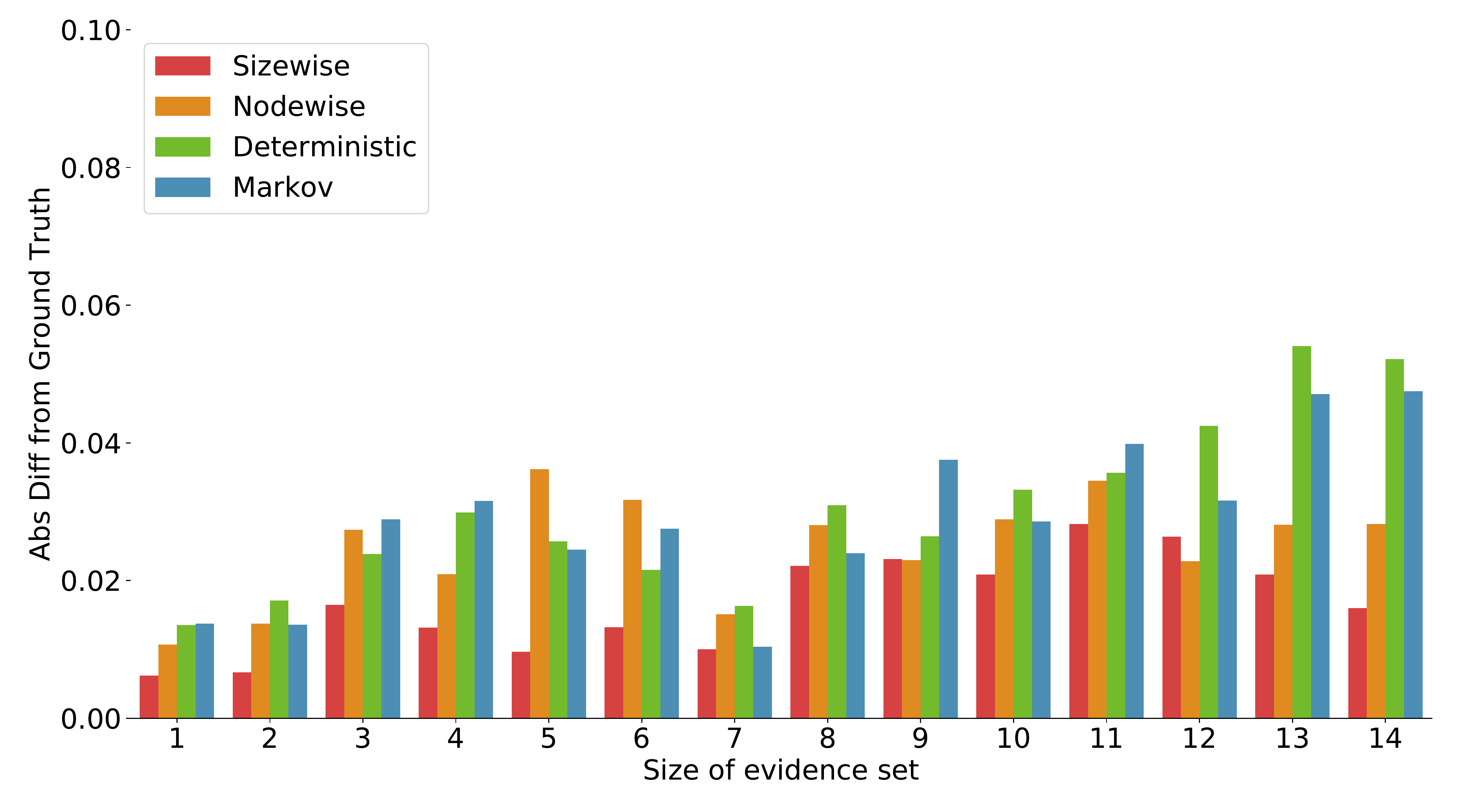}}
    \subfigure[Markov evidence
    \label{app:fig:markov_masking_D1_short}
    ]{\includegraphics[width=0.45\textwidth]{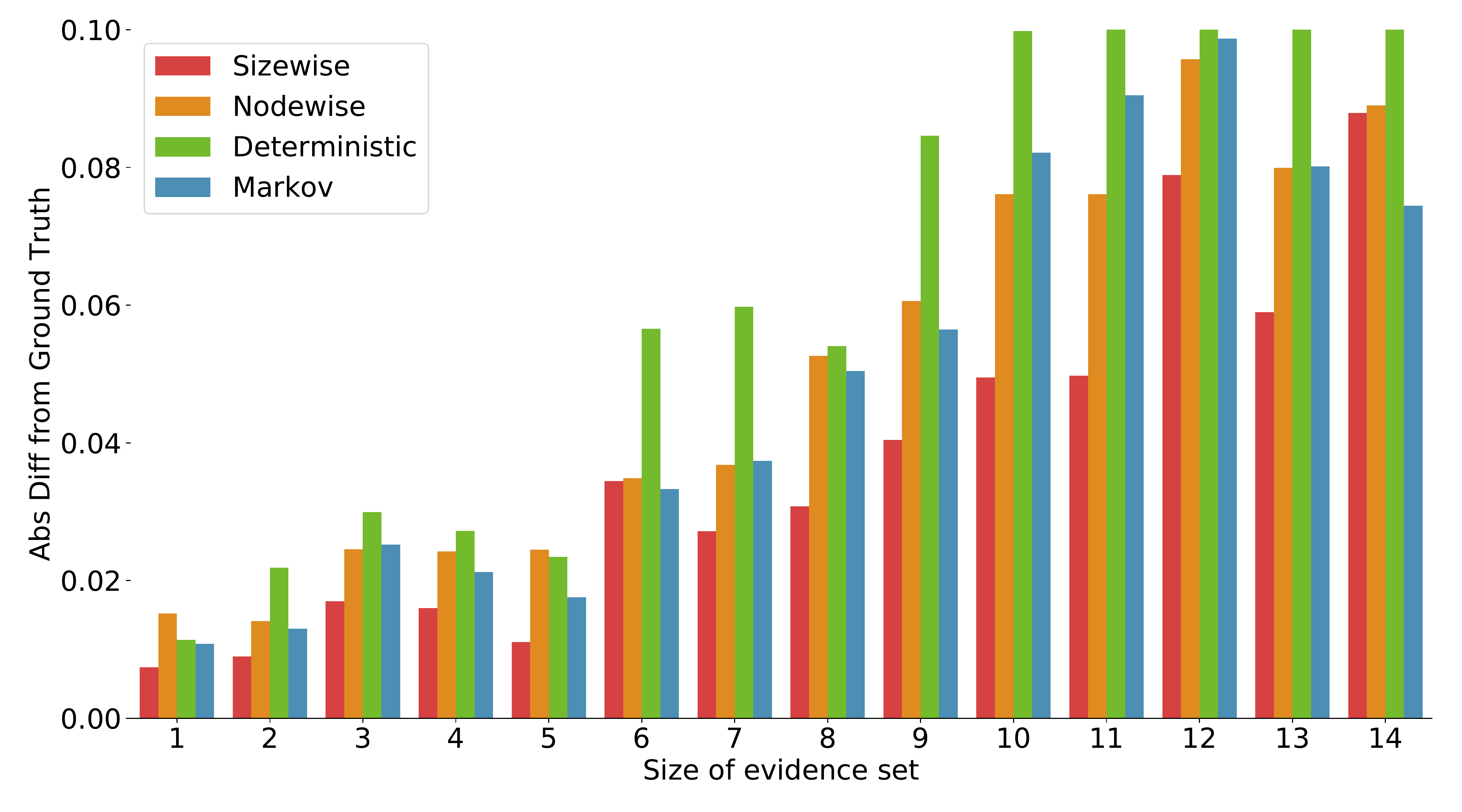}}
    \caption{\label{app:fig:D1_short_trained_error_plot}
    Models trained for 20 epochs exhibit poorer performance on average and the effect of different masking schemes is more pronounced. One epoch corresponds to the UM using 10 million samples. Note the different scale of the y-axis compared to Figures~\ref{fig:error_masking_plot} and \ref{app:fig:D2_full_trained_error_plot}.}
\end{figure}
\begin{figure}
    \centering
    \subfigure[Uniform evidence \label{app:fig:uni_masking_D1_short_slope} ]{\includegraphics[width=0.45\textwidth]{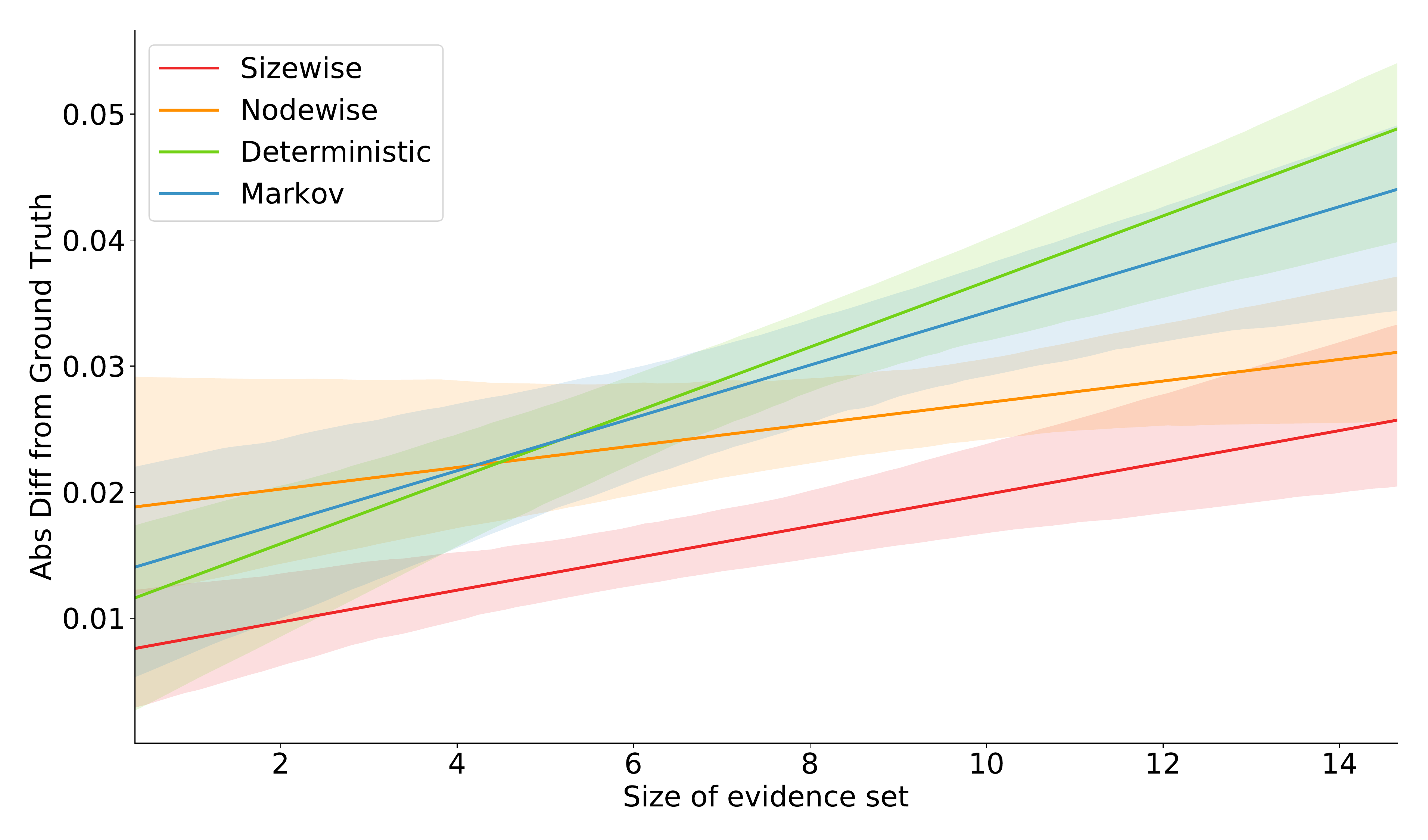}}
    \subfigure[Markov evidence
    \label{app:fig:markov_masking_D1_short_slope}
    ]{\includegraphics[width=0.45\textwidth]{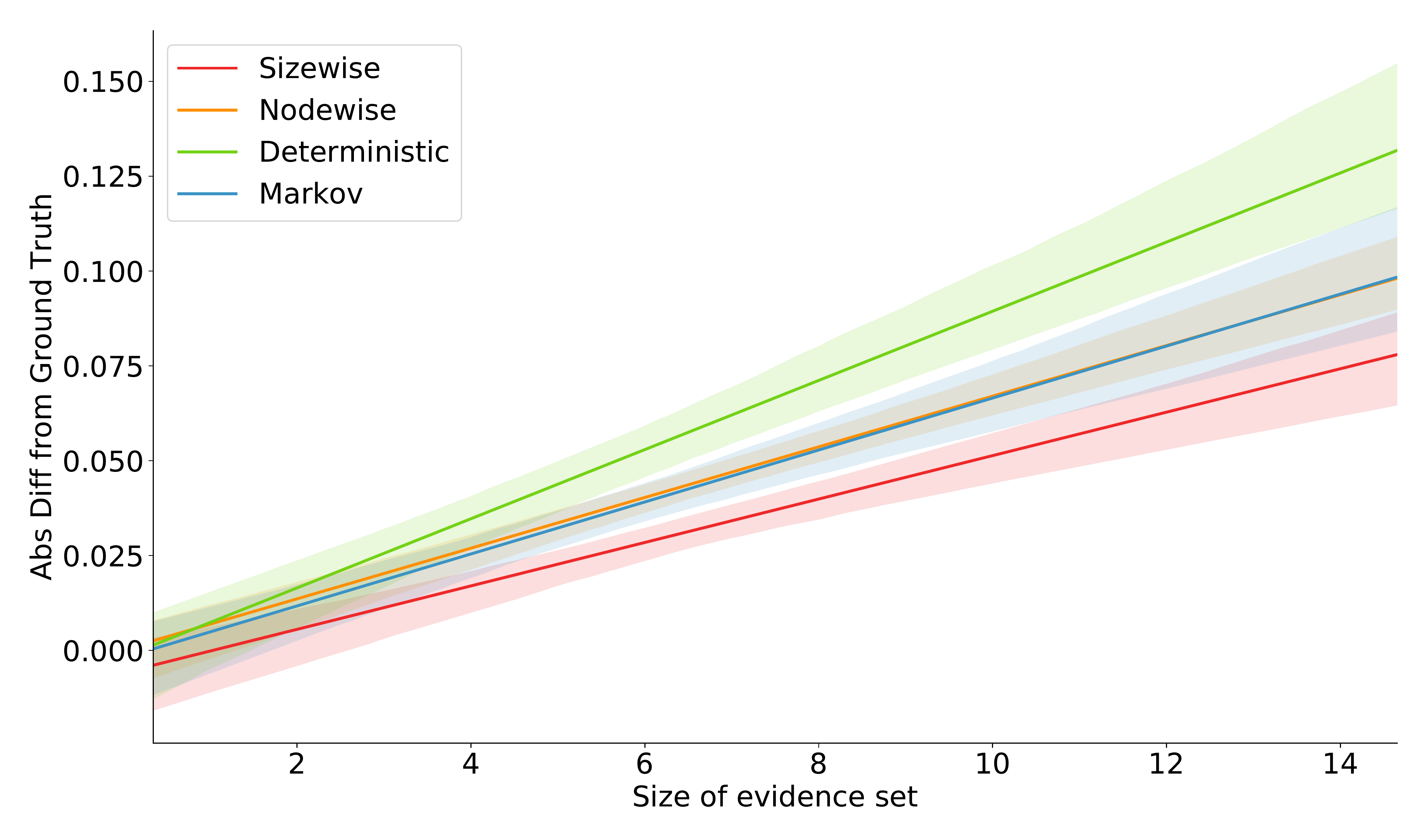}}
    \caption{\label{app:fig:D1_short_trained_error_plot_slope}
    Linear model plots corresponding to the same experiment as Figure~\ref{app:fig:D1_short_trained_error_plot}. For a small number of epochs, the sizewise masking distribution is the best scheme to use for both the uniform and the Markov evidence.}
\end{figure}

During training with the different masking distributions, the UM obtained more than $10^{11}$ possible states of the $24$-node BN during training, which is nearly four orders of magnitude larger than the size of the state space.
In larger BNs with hundreds of nodes, the UM would not be able to get samples from substantial parts of the state space during training.
In addition, for the masking distributions that satisfy $M(b_i)\in \{0,1\}$ for all $i$, it is expected that a denoising-autoencoder will behave as a UM after obtaining a sufficiently large number of samples, making comparisons between masking schemes difficult.
UMs trained with uniform masking exhibit good performance~\citep{Douetal17} even for large BNs.

In order to simulate the scenario of training a large BN with the same small sythentic BN available, we trained another series of models for a much smaller number of epochs, in order to see if differences between the masking schemes are more pronounced. Figures~\ref{app:fig:D1_short_trained_error_plot} and \ref{app:fig:D1_short_trained_error_plot_slope} correspond to this experiment.
Across the various queries considered during this experiment (most of which are not shown), we noticed that the deterministic cycle scheme had the worst behaviour whereas the sizewise scheme was the most performant.

\subsection{Masking schemes with iid data}

\begin{table}[h!]
    \centering
    \begin{tabular}{|c|c|c|}
\hline
Feature &MCAR-trained &  Sizewise-trained\\
\hline
0	&0.02748	(0.044)	&0.02838	(0.046)\\
1	&0.02160	(0.037)	&0.02207	(0.038)\\
2	&0.00936	(0.017)	&0.00942	(0.017)\\
3	&0.02511	(0.050)	&0.02534	(0.051)\\
4	&0.00386	(0.031)	&0.00389	(0.031)\\
5	&0.00856	(0.073)	&0.00845	(0.073)\\
6	&0.00404	(0.012)	&0.00406	(0.012)\\
7	&0.01597	(0.046)	&0.01584	(0.046)\\
\hline
    \end{tabular}
    \caption{MCAR corrupted data. Mean squared error of reconstruction per feature and standard deviation.}
    \label{tab:MCAR}
\end{table}


\begin{table}[h!]
    \centering
    \begin{tabular}{|c|c|c|}
\hline
Feature &MCAR-trained &Sizewise-trained\\
\hline
0 &0.03181	(0.051)	&0.03202	(0.051)\\
1 &0.02286	(0.038)	&0.02317	(0.038)\\
2 &0.00980	(0.017)	&0.01002	(0.017)\\
3 &0.02619	(0.052)	&0.02617	(0.051)\\
4 &0.00370	(0.030)	&0.00372	(0.030)\\
5 &0.00878	(0.075)	&0.00878	(0.075)\\
6 &0.00410	(0.011)	&0.00411	(0.011)\\
7 &0.01539	(0.041)	&0.01524	(0.042)\\
\hline
    \end{tabular}
    \caption{Structured-corrupted data. Mean squared error of reconstruction per feature and standard deviation.}
    \label{tab:str-corr}
\end{table}

In this section, we included an experiment with the variational autoencoder with arbritrary conditioning (VAEAC) architecture~\citep{Ivaetal19}, together with the yeast dataset \citep{Dua:2019}. \citet{Ivaetal19} consider a single corruption distribution called "Missing completely at random" (MCAR) which corrupts the original data by randomly turning observed values into unobserved. The masking scheme MCAR with parameter equal to 0.5 corresponds to what our masking scheme called \textbf{uniform power setwise masking distribution}; see Section~\ref{sec:masks} for details. Their experimental setup goes as follows: a VAE is first trained to reconstruct the corrupted dataset with the use of another MCAR distribution that is different from the one used to corrupt the dataset. Since the VAE can utilize any reasonable masking distribution in order to recover the original values of the features, we extended this
experiment. Specifically,  the dataset was corrupted and reconstructed using different masking distributions for each step. Then, the mean-squared errors per feature were compared for each configuration of masking and corruption processes. In Table~\ref{tab:MCAR}, the data were corrupted by an MCAR with probability of observing a single feature $p=0.5$ per row of the dataset. The VAE that uses a sizewise distribution during training does not have a considerable advantage over the MCAR case.

We also picked the corruption distribution to be different from the possible masking schemes used for training. In Table~\ref{tab:str-corr}, we consider a corruption distribution that induces some structure - we call this a structured-corrupted distrbution (SC). The corruption scheme SC always masks neighboring triplets of features, i.e., $(0,1,2)$, $(4,5,6)$, etc. as opposed to MCAR, which independently corrupts different features. The SC corruption distribution emulates situations where only a subset of relevant features is observed at any given time. As expected, the VAEAC using MCAR or uniform-sizewise during training exhibits larger error, on average, when the ground truth masking is SC.  
We conjecture that part of this error is because the VAE cannot capture couplings between the predicted distributions given evidence but further investigations are needed to verify such claim.

\end{document}